\documentclass{article}

\usepackage[preprint]{neurips_2026}

\usepackage[utf8]{inputenc}
\usepackage[T1]{fontenc}
\usepackage{microtype}
\usepackage{graphicx}
\usepackage{booktabs}
\usepackage{amsmath, amssymb}
\usepackage{xcolor}
\usepackage{hyperref}
\usepackage{url}
\usepackage{algorithm}
\usepackage{algpseudocode}
\usepackage{multirow}
\usepackage{array}
\usepackage{caption}
\usepackage{subcaption}
\usepackage{enumitem}
\usepackage{tabularx}
\usepackage{tikz}
\usepackage{bbm}
\usetikzlibrary{positioning, arrows.meta, fit, backgrounds, shapes.geometric, calc, decorations.pathreplacing}
\usepackage{booktabs}
\usepackage{tabularx}
\usepackage{multirow}
\usepackage{siunitx}

\hypersetup{
  colorlinks=true,
  linkcolor=black,
  citecolor=blue!60!black,
  urlcolor=blue!60!black,
}

\setlist{nosep}

\title{Structural Interpretations of Protein Language Model
Representations via Differentiable Graph Partitioning}

\author{
  Siddhant Dutta\thanks{Equal contribution.} \and
  Edward Tan Beng Wai\footnotemark[1] \and
  Soumick Sarker\footnotemark[1] \and
  Pasan Gunawardane\footnotemark[1] \and
  Jagath C. Rajapakse \\
  Nanyang Technological University, Singapore \\
  \texttt{\{siddhant010, soumick001, ed0001ai, c250135\}@e.ntu.edu.sg} \\
  \texttt{ASJagath@ntu.edu.sg}
}

\begin{document}

\maketitle

\begin{abstract}
    Protein language models such as ESM-2 learn rich residue representations
that achieve strong performance on protein function prediction, but their
features remain difficult to interpret as structural \& evolutionary
signals are encoded in dense latent spaces. We propose a plug-\&-play framework that
projects ESM-2 representations onto protein contact graphs \& applies
\textbf{SoftBlobGIN}, a lightweight Graph Isomorphism Network with
differentiable Gumbel-softmax substructure pooling, to perform
structure-aware message passing \& learn coarse functional substructures
for downstream prediction tasks. Across enzyme classification, SoftBlobGIN achieves 92.8\%
accuracy \& 0.898 macro-F1. Unlike post hoc analysis of protein
language models alone, our method produces directly auditable structural
explanations: GNNExplainer recovers biologically meaningful active-site
residues, spatially localized functional clusters, \& catalytic contact
patterns. On binding-site detection, SoftBlobGIN improves residue AUROC
from $0.885$ using an ESM-2 linear probe to $0.983$, indicating that these
structural explanations are not recoverable from language-model features
alone. Learned blob partitions provide an additional layer of
interpretability by automatically grouping residues into functional
substructures, with blobs containing annotated active-site residues showing
$1.85\times$ higher importance than other blobs ($\rho{=}0.339$,
$p{=}0.009$), without any active-site supervision. Our framework requires no retraining of the language model, adds only
$\sim$1.1M parameters, \& generalises across ProteinShake tasks, achieving
$F_{\max}$ of $0.733$ on Gene Ontology prediction \& AUROC of $0.969$ on
binding-site detection. We position this as an interpretable structural
companion to protein language models that makes their predictions more
transparent \& auditable.
\end{abstract}

\section{Introduction}
\label{sec:intro}


Despite the explosion of
sequence databases, a large fraction of sequenced proteins remain
functionally unannotated \citep{kustatscher2022understudied}. 
The rise of large-scale structural data, such as AlphaFold~2 \citep{jumper2021highly} provides predicted structures for over 200 million
proteins. Recently, Protein Language Models (PLMs) such as ESM-2 \citep{lin2023evolutionary}
have transformed function prediction. ESM-2 in particular has become a
\emph{de facto} feature extractor: simple multilayer perceptrons (MLPs) on its mean-pooled embeddings
already match or exceed structure-based graph neural networks (GNNs) on many tasks. This success,
however, comes at the cost of \emph{interpretability}. ESM-2 embeddings are
1280-dimensional dense vectors with no obvious mapping to specific residues,
contacts, or biochemical motifs.
However, many downstream tasks need interpretability, specifically to deploy in clinical settings where safety \& regulatory compliance are paramount. Furthermore, interpretable models allow researchers to verify that computational predictions correspond to meaningful biological mechanisms as opposed to spurious data correlations, potentially discovering novel insights. 


Existing language-model probes (linear classifiers on attention heads, attention rollout, etc.) recover
broad sequence patterns but rarely surface spatially localised, biochemically
specific motifs. On the other hand, structural methods such as conventional GNNs on protein contact graphs often use fixed-radius neighbourhoods, limiting flexibility to empirically defined constants. This means a catalytic residue forming a 3\,\AA{} hydrogen bond with
its substrate \& a surface residue 7.9\,\AA{} from a distant loop receive
identical graph topology. Recent works \citep{wang2025bioblobs} address this by using variable-size partitions, based on a Geometric Vector Perceptron (GVP) encoder \& a vector-quantized (VQ) codebook, though at the cost of greater computation, \& an interpretability gap. Importantly, these methods do not leverage the expressivity of recent PLM representations.

This raises the question of \textit{when} structural reasoning adds information beyond what PLMs already capture. Our empirical answer is that the boundary is biological. For graph-level function tasks like enzyme classification (EC), ESM-2 mean-pooling is nearly sufficient \& graph structure adds little insights. For residue-level structural tasks like binding-site detection, message-passing over the contact graph adds substantial information that ESM-2 alone cannot recover. The interesting consequence is interpretability: in the regime where structure matters, we want models whose structural reasoning is auditable. We therefore propose a computationally lightweight, structurally interpretable GNN, which jointly leverages semantically rich ESM-2 features, while remaining interpretable. To that end, we list our contributions:


\begin{enumerate}[leftmargin=1.3em,itemsep=2pt,topsep=2pt]
  \item \textbf{Empirical characterisation.} We map when structural
        reasoning helps frozen PLM features. For graph-level EC task, ESM-2 mean-pooling is nearly sufficient
        ($0.910$ vs $0.912$ accuracy with vs without the contact graph).
        For residue-level binding-site detection, message-passing over
        the contact graph closes a $9.8$-point AUROC gap that ESM-2
        alone cannot.
  \item \textbf{Interpretable architecture.} We introduce \textbf{SoftBlobGIN}
        that replaces BioBlobs' GVP encoder \& VQ codebook with a single
        Gumbel-softmax assignment head \citep{jang2017gumbel}, producing
        $K$ differentiable, soft protein substructures with ${\sim}1.1$M
        parameters \& no language-model retraining.
  \item \textbf{Biological validation.} We quantitatively validate the
        resulting explanations against established enzyme biochemistry.
        GNNExplainer \citep{ying2019gnnexplainer} recovers
        catalytic-residue enrichment, active-site burial, spatial
        co-localisation, \& tertiary-contact geometry consistent with
        catalytic triads. Learned blobs spontaneously separate functional
        sites from structural scaffold, with active-site-containing blobs
        carrying $1.85\times$ higher importance ($\rho{=}0.339$,
        $p{=}0.009$).
\end{enumerate}

\section{Related Work}
\label{sec:related}

\paragraph{Graph neural networks for protein structure.}
Protein structures are naturally represented as residue contact graphs where
nodes correspond to amino acids \& edges encode spatial proximity or geometric
relations. Conventional GNNs such as Graph Convolutional
Networks (GCN) \citep{kipf2017gcn} \& Graph Attention Networks (GAT)
\citep{velickovic2018gat} have therefore been widely adopted for structure-based
protein learning. However, these architectures are limited in expressive power.
\citet{xu2019gin} showed that both GCN \& GAT are strictly less expressive than
the Weisfeiler--Leman (WL) graph isomorphism test, motivating the introduction
of Graph Isomorphism Networks (GIN), which match the WL upper bound.
Subsequent work extended GIN to incorporate edge information through GINEConv
\citep{hu2019strategies}, enabling explicit modeling of pairwise residue geometry
such as C\textsubscript{$\alpha$}--C\textsubscript{$\alpha$} distances. These
developments make GIN-style architectures a natural foundation for protein
graphs, where local geometric interactions are often functionally informative.

\paragraph{Protein language models (PLMs).}
In parallel, PLMs have substantially advanced function
prediction by learning sequence-derived representations at scale. Models such as
ESM-2 \citep{lin2023evolutionary} are trained on approximately 65 million
protein sequences by using masked language modeling \& produce rich per-residue
embeddings encoding evolutionary conservation, structural regularities, \&
functional context. In many downstream settings, frozen ESM-2 features combined
with lightweight classifiers already achieve competitive or state-of-the-art
performance. This has established PLM as strong general
feature extractors, but has also introduced an interpretability challenge.
Because these representations are high-dimensional dense vectors, the structural
or biochemical signals responsible for a prediction are not directly observable.
Our work builds on this observation by treating ESM-2 as a frozen semantic
encoder while introducing explicit graph-based structural reasoning.

\paragraph{Hierarchical pooling \& structural abstraction.}
Beyond residue-level message passing, hierarchical pooling methods aim to learn
coarse structural abstractions by grouping residues into higher-level
substructures. DiffPool \citep{ying2018diffpool} introduced differentiable soft
cluster assignments for hierarchical graph coarsening, enabling end-to-end
learning of graph hierarchies. More recently, BioBlobs
\citep{wang2025bioblobs} adapted this paradigm to proteins through biologically
motivated blob partitions, demonstrating strong performance on ProteinShake
benchmarks. However, this approach relies on a substantially heavier architecture
involving Geometric Vector Perceptrons \& vector quantization. In contrast, our
SoftBlobGIN replaces these components with lightweight Gumbel-softmax pooling
\citep{jang2017gumbel}, allowing differentiable learning of soft functional
substructures with substantially lower computational overhead.

\paragraph{Explainability for graph neural networks.}
Interpreting GNN predictions has motivated a growing body of
post hoc explanation methods. GNNExplainer \citep{ying2019gnnexplainer} learns
continuous edge \& feature masks by maximizing mutual information between
selected subgraphs \& model predictions, providing instance-specific
substructure explanations. Integrated Gradients
\citep{sundararajan2017axiomatic} offers a complementary attribution framework
by integrating gradients along a path from a baseline input to the observed
example. These methods have become standard tools for probing graph models, with
evaluation typically based on metrics such as fidelity, sparsity, \&
characterization \citep{pope2019gnnexp,yuan2022explainability}. In this work,
we employ both methods to evaluate whether structure-aware reasoning over frozen
PLM embeddings produces biologically meaningful \& auditable
explanations.

\section{Problem \& Evaluation Criteria}
\label{sec:problem}

We aim to learn a function classifier $f_\theta$ over protein contact 
graphs that produces sparse, biologically faithful explanations for its 
predictions. An explanation is a pair of continuous masks $(M, F)$ over 
edges \& node features. We evaluate explanations along two axes:

\paragraph{Predictive faithfulness.} Standard fidelity-based metrics 
applied to graph explanations: sparsity, sufficiency 
(Fid\textsuperscript{+}), necessity (Fid\textsuperscript{-}), \& 
intra-class feature-mask stability (definitions in 
Appendix~\ref{app:met}).

\paragraph{Biological faithfulness.} Predictive metrics are necessary 
but insufficient: an explanation can be faithful to the model \& 
still biologically meaningless. We additionally require explanations 
to align with established enzyme biochemistry along four axes (B1)-(B4):

\paragraph{(B1) Catalytic-residue enrichment.} For each EC class $c$ \&
amino acid $a \in \Sigma$, let
$\hat{p}_{c,a}^{\mathrm{top}}$ \& $\hat{p}_{c,a}^{\mathrm{bg}}$ be the
empirical frequencies of $a$ in $\mathcal{I}_{0.20}(G)$ \& $V$, respectively.
The log-enrichment is
\begin{equation}
\mathrm{Enr}(c,a) \;=\; \log_2 \frac{\hat{p}_{c,a}^{\mathrm{top}} + \delta}{\hat{p}_{c,a}^{\mathrm{bg}} + \delta}, \qquad \delta = 10^{-6}.
\label{eq:enrichment}
\end{equation}
We test the hypothesis
$\mathbb{E}_a\!\big[\mathrm{Enr}(c,a) \mid a \in \mathrm{Cat}\big] >
\mathbb{E}_a\!\big[\mathrm{Enr}(c,a) \mid a \notin \mathrm{Cat}\big]$
where $\mathrm{Cat} = \{\mathrm{H, C, S, D, E, K, R, Y}\}$.

\paragraph{(B2) Active-site burial.} The expected SASA gap between
important \& unimportant residues:
\begin{equation}
\Delta_{\mathrm{SASA}}(c) \;=\;
\mathbb{E}_{i \in \mathcal{I}_{0.20}(G)}\!\big[\mathrm{SASA}_i\big]
\;-\;
\mathbb{E}_{i \notin \mathcal{I}_{0.20}(G)}\!\big[\mathrm{SASA}_i\big]
\;<\; 0.
\label{eq:sasa-gap}
\end{equation}

\paragraph{(B3) Spatial co-localisation.} Let
$\bar{D}(\mathcal{I}) = \binom{|\mathcal{I}|}{2}^{-1}\sum_{i<j \in \mathcal{I}} \|c_i - c_j\|_2$
be the mean pairwise distance. Sampling $B=100$ random subsets
$\mathcal{R}^{(b)} \subset V$ with $|\mathcal{R}^{(b)}| = |\mathcal{I}_{0.20}(G)|$,
the spatial-clustering $z$-score is
\begin{equation}
Z_{\mathrm{spatial}}(G) \;=\; \frac{\bar{D}\big(\mathcal{I}_{0.20}(G)\big) - \mu_R}{\sigma_R},
\quad
\mu_R = \tfrac{1}{B}\sum_b \bar{D}(\mathcal{R}^{(b)}),\;
\sigma_R = \mathrm{std}_b\!\big[\bar{D}(\mathcal{R}^{(b)})\big].
\label{eq:zspatial}
\end{equation}
We require $\mathbb{E}_G[Z_{\mathrm{spatial}}(G)] < 0$ (more compact than
random).

\paragraph{(B4) Tertiary-contact preference.} Among edges with $M_e \ge 0.5$,
the fraction with sequence separation $|i\!-\!j| > 20$ should exceed that of
the unimportant set, \& the C\textsubscript{$\alpha$}--C\textsubscript{$\alpha$}
distance distribution should peak in the catalytic-triad regime
$d_{ij} \in [6, 10]\,$\AA{}.

\medskip

The remainder of the paper presents architectures
(Section~\ref{sec:method}) that parameterise $f_\theta$ at varying levels of
structural expressivity, results on the
$\mathcal{L}_{\mathrm{cls}}$-objective (Section~\ref{sec:experiments}), \&
joint validation against both fidelity (Section~\ref{sec:explanations}) \&
the biological criteria (B1)--(B4) which we report in Section~\ref{subsec:coref}, all with the goal of obtaining a model that is simultaneously \emph{accurate} \& \emph{auditable}.

\section{Method}
\label{sec:method}
 
\begin{figure*}[ht]
\centering
\resizebox{\textwidth}{!}{%
\begin{tikzpicture}[
    >=stealth,
    mainblock/.style={rectangle, draw, thick, align=center, rounded corners=4pt, minimum height=1.8cm, minimum width=3.5cm, font=\sffamily\bfseries},
    panel/.style={rectangle, draw, dashed, thick, rounded corners=8pt, inner sep=6pt, fill=gray!5}, 
    paneltbl/.style={font=\sffamily\bfseries, anchor=south},
    mathnode/.style={font=\small, align=center},
    arrow/.style={->, thick, >=stealth},
    esmcolor/.style={fill=purple!15},       
    blobcolor/.style={fill=cyan!15},        
    interpcolor/.style={fill=orange!15},    
    graphcolor/.style={fill=gray!15}        
]


\node (seq) [align=center, font=\sffamily\small] {Protein Sequence\\$S \in \Sigma^N$};
\node (esm) [mainblock, esmcolor, right=0.8cm of seq] {\textbf{1. ESM-2 PLM}\\Dense Embeddings};

\node (struct) [below=1.5cm of seq, align=center, font=\sffamily\small] {3D Coordinates\\$C \in \mathbb{R}^{N \times 3}$};
\node (graphbuild) [mainblock, graphcolor, minimum width=2.4cm, minimum height=1.2cm, right=1.0cm of struct] {Radius Graph\\$\varepsilon=8$\AA};

\node (gine) [mainblock, graphcolor, right=3.2cm of esm, yshift=-0.9cm] {GINEConv\\Backbone};

\node (blob) [mainblock, blobcolor, right=3.2cm of gine] {\textbf{2. SoftBlobGIN}\\Differentiable GS};
\node (explain) [mainblock, interpcolor, right=3.2cm of blob] {\textbf{3. Interpretable}\\Structure \& EC};

\draw[arrow] (seq.east) -- (esm.west);
\draw[arrow] (struct.east) -- (graphbuild.west);

\draw[arrow] (esm.east) to[out=0, in=180] node[above, sloped, font=\small, pos=0.55] {$\mathbf{X} \in \mathbb{R}^{1318}$} (gine.west);
\draw[arrow] (graphbuild.east) to[out=0, in=180] node[below, sloped, font=\small, pos=0.55] {$E, \mathbf{E}_{\mathrm{attr}}$} (gine.west);

\draw[arrow] (gine.east) -- node[above, font=\small] {$H^{(L)}$} (blob.west);
\draw[arrow] (blob.east) -- node[above, font=\small] {$\{b_k\}_{k=1}^K$} (explain.west);


\node (pa_center) [below=3.4cm of gine] {};
\node (pa_eq1) [mathnode, esmcolor, draw, rounded corners, inner sep=4pt] at (pa_center) {
    \textbf{Node Features (1318-d)}\\
    $x_i = [\,\mathbf{\phi^{\mathrm{esm}}} \,\|\, \phi^{\mathrm{phys}} \,\|\, \phi^{\mathrm{sasa}} \,\|\, \dots \,]$
};
\node (pa_eq2) [mathnode, below=0.4cm of pa_eq1] {
    \textbf{GINE Message Passing}\\
    $h_i^{(\ell)} = \mathrm{MLP}\Big((1 \!+\! \epsilon) h_i^{(\ell-1)} + \sum_{j \in \mathcal{N}(i)} \dots\Big)$
};

\node (pb_center) [below=3.4cm of blob] {};
\node (pb_eq1) [mathnode, blobcolor, draw, rounded corners, inner sep=4pt] at (pb_center) {
    \textbf{Gumbel-Softmax Assignment}\\
    $A_{ik} = \frac{\exp((L_{ik} + g_{ik})/\tau_t)}{\sum_{k'=1}^K \exp((L_{ik'} + g_{ik'})/\tau_t)}$
};
\node (pb_eq2) [mathnode, below=0.4cm of pb_eq1] {
    \textbf{Learned Blob Partitions}\\
    $b_k = g_\psi\!\left(\mathrm{LN}\!\left( \frac{\sum_i A_{ik} h_i^{(L)}}{\sum_i A_{ik} + \epsilon} \right)\right)$
};

\node (pc_center) [below=3.4cm of explain] {};
\node (pc_eq1) [mathnode] at (pc_center) {
    \textbf{Dual Graph Embedding}\\
    $z(G) = \big[\, \max_{k} b_k \,\|\, \frac{1}{N}\sum_i h_i^{(L)} \big]$
};
\node (pc_eq2) [mathnode, interpcolor, draw, rounded corners, inner sep=4pt, below=0.4cm of pc_eq1] {
    \textbf{Explanation Objective}\\
    $\min_{M, F} -\log [f_\theta(G_M)]_{\hat{y}} + \lambda_1\|M\|_1 \dots$\\
    {\footnotesize (Recovers Active Sites \& Catalytic Triads)}
};

\begin{scope}[on background layer]
    \node[panel, fit=(pa_eq1) (pa_eq2), minimum width=5.2cm, minimum height=3.5cm] (panel_a) {};
    \node[panel, fit=(pb_eq1) (pb_eq2), minimum width=5.4cm, minimum height=3.5cm] (panel_b) {};
    \node[panel, fit=(pc_eq1) (pc_eq2), minimum width=5.4cm, minimum height=3.5cm] (panel_c) {};
\end{scope}

\node[paneltbl, above=0.1cm of panel_a.north] {(a) PLM-to-Graph Projection};
\node[paneltbl, above=0.1cm of panel_b.north] {(b) Differentiable Blob Partitioning};
\node[paneltbl, above=0.1cm of panel_c.north] {(c) Interpretable Structural Readout};

\draw[thick, dotted, gray] (gine.south) -- (panel_a.north);
\draw[thick, dotted, gray] (blob.south) -- (panel_b.north);
\draw[thick, dotted, gray] (explain.south) -- (panel_c.north);

\end{tikzpicture}
}
\caption{\textbf{Overview of the SoftBlobGIN Framework.} Our pipeline acts as an interpretable structural companion to protein language models. \textbf{(a)} Dense, opaque ESM-2 representations ($\phi^{\mathrm{esm}}$) are concatenated with explicit structural/physicochemical features and projected onto a 3D contact graph. \textbf{(b)} A lightweight, differentiable Gumbel-Softmax (GS) pooling head learns to softly partition residues into functional substructures (blobs). \textbf{(c)} The resulting dual-readout graph embedding enables high-accuracy EC classification while seamlessly supporting post-hoc attribution methods to extract biologically faithful motifs, such as catalytic triads and buried active sites.}
\label{fig:pipeline}
\end{figure*}
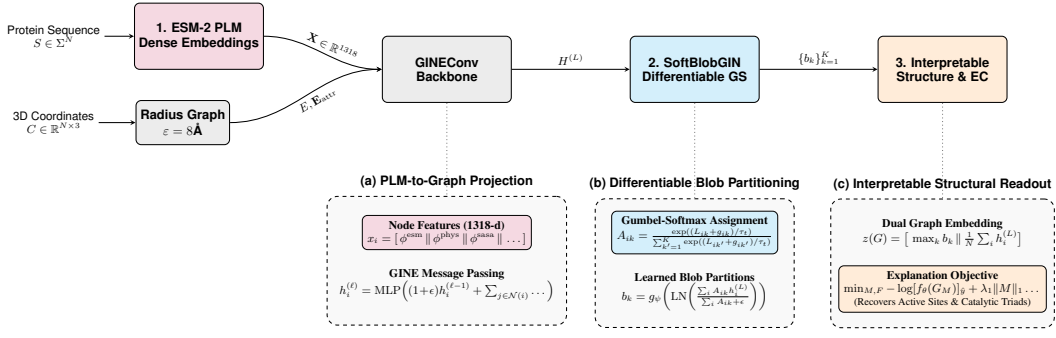  
 
\subsection{Architecture}
\label{sec:architectures}
 
Our model has three components: (i)~a GIN backbone that performs
message-passing over the protein contact graph, (ii)~a Gumbel-softmax
blob-pooling head that partitions residues into $K$ learned substructures,
\& (iii)~a classifier that reads out from both blob-level \& global
representations. We describe each below; full hyperparameters \& baseline
architectures are in Appendix~\ref{app:base}.
 
\paragraph{GIN backbone.}
Each residue's 1318-d feature vector (Appendix~\ref{app:feat}) is projected
to a hidden dimension $\hbar = 256$ via a linear layer. We then apply $L=4$
GINEConv layers \citep{hu2019strategies}, each parameterised by a 2-layer
MLP with BatchNorm \& ReLU. Edge features ($d_e = 18$, encoding
radial-basis-expanded C\textsubscript{$\alpha$} distances \& sequence
separation; see Appendix~\ref{app:feat}) are injected at every layer. After
$L$ layers, each residue carries a $\hbar$-dimensional representation
$h_i^{(L)}$ that integrates information from its local structural
neighbourhood.
 
\paragraph{Differentiable blob pooling (SoftBlobGIN).}
To obtain an interpretable graph-level representation, we partition residues
into $K{=}8$ soft, learned substructures via Gumbel-softmax assignment
\citep{jang2017gumbel}. An MLP head
$f_\phi: \mathbb{R}^{\hbar} \to \mathbb{R}^K$ produces assignment logits
$L_{ik}$ for each residue $i$ \& blob $k$. With Gumbel noise
$g_{ik} \stackrel{\text{iid}}{\sim} \mathrm{Gumbel}(0,1)$ \& temperature
$\tau_t$ annealed linearly from $1.0$ to $0.1$ across training:
\begin{equation}
A_{ik}(\tau_t) \;=\; \frac{\exp\big((L_{ik} + g_{ik})/\tau_t\big)}
                          {\sum_{k'=1}^{K} \exp\big((L_{ik'} + g_{ik'})/\tau_t\big)},
\qquad \sum_{k} A_{ik} = 1.
\label{eq:gumbel}
\end{equation}
Each blob embedding is the assignment-weighted mean of node representations,
refined by a 2-layer MLP $g_\psi$ with LayerNorm:
\begin{equation}
b_k \;=\; g_\psi\!\Bigg(\mathrm{LN}\!\bigg(
   \frac{\sum_{i=1}^{N} A_{ik}\, h_i^{(L)}}
        {\sum_{i=1}^{N} A_{ik} + \epsilon}
\bigg)\Bigg) \in \mathbb{R}^{\hbar}, \quad k = 1,\dots,K.
\label{eq:blob-readout}
\end{equation}
This replaces BioBlobs' GVP encoder \& VQ codebook
\citep{wang2025bioblobs} with a single MLP assignment head, reducing the
pooling module to ${\sim}$35K parameters while retaining differentiable,
non-overlapping substructure discovery.
 
\paragraph{Readout \& classifier.}
The graph embedding concatenates a blob-level max-pool with a global
mean-pool:
\begin{equation}
z(G) \;=\; \big[\, \max_{k \in [K]} b_k \;\|\;
\tfrac{1}{N}\textstyle\sum_i h_i^{(L)} \,\big]
\;\in \mathbb{R}^{2\hbar}.
\label{eq:readout}
\end{equation}
A 2-layer MLP with BatchNorm maps $z(G)$ to class logits. The full model
has ${\sim}$1.1M trainable parameters; Algorithm~\ref{alg:sbgin}
(Appendix~\ref{app:alg}) gives the complete forward pass.
 
\paragraph{Baselines.}
To isolate the contribution of graph structure, we compare against two
non-graph baselines (Seq MLP on amino-acid composition; Residue MLP on
mean-pooled 1318-d features) \& a plain GIN without blob pooling, which
instead uses JumpingKnowledge concatenation \citep{xu2018jumping} with dual
mean--max readout (${\sim}$1.4M parameters). All share the same feature
pipeline. Details are in Appendix~\ref{app:base}.
 
\paragraph{Task-specific heads.}
For graph-level tasks (EC classification, Gene Ontology, Protein Family),
we use SoftBlobGIN with the blob-pooling readout above. For node-level
tasks (binding-site detection), blob pooling is inapplicable because it
produces graph-level rather than residue-level representations; we
therefore use the GIN backbone directly with a per-residue MLP head on
the \textit{JumpingKnowledge} output. We note this distinction explicitly: results
on node-level tasks reflect the GIN backbone's message-passing, not the
blob-pooling module. For pairwise tasks (structure similarity, PPI), we
use Siamese or bilinear heads on the SoftBlobGIN encoder
(Appendix~\ref{app:base}).

\subsection{Explainability \& training}
\label{sec:explainability}

We apply the explanation framework to the trained GIN classifier using both
GNNExplainer \& Integrated Gradients (IG). For GNNExplainer, we optimise
continuous edge \& feature masks
\begin{equation}
(M^*, F^*) = \arg\min_{M,F}
-\log \big[f_\theta(G_M)\big]_{\hat{y}}
+ \lambda_1 \frac{\|M\|_1}{|E|}
+ \lambda_2 \mathcal{H}(M)
+ \lambda_3 \frac{\|F\|_1}{d}
+ \lambda_4 \mathcal{H}(F),
\label{eq:gnnexp-objective}
\end{equation}
where $G_M$ denotes the masked graph \& $\mathcal{H}(\cdot)$ is the
element-wise entropy regulariser. Integrated Gradients computes attribution scores by accumulating gradients
along a linear path from baseline input $x'$ to input $x$:
\begin{equation}
\mathrm{IG}_i(x)
=
(x_i - x_i')
\int_{\alpha=0}^{1}
\frac{\partial f(x' + \alpha(x-x'))}{\partial x_i}
\, d\alpha,
\label{eq:ig}
\end{equation}
approximated numerically with a 50-step trapezoidal rule. Per-protein GNNExplainer optimisation runs for 300 gradient steps, after which we evaluate explanations using fidelity, stability, unfaithfulness, \& 
characterization metrics. Stability quantifies robustness to perturbations.

\paragraph{Class-imbalanced training.} We minimise $\mathcal{L}_{\mathrm{cls}}$
(Eq.~\ref{eq:focal}) with AdamW (lr $=10^{-3}$, weight decay $=10^{-4}$),
cosine annealing with 10-epoch warmup, gradient clipping at norm 1.0, edge
dropout (5\%) \& feature masking (5\%) during training, up to 200 epochs
with patience 30 on val loss. For the ensemble, five SoftBlobGIN models
with different seeds are averaged at the softmax level. Because the class frequencies
$\pi_c = \mathbb{P}(Y\!=\!c)$ are highly skewed
($\pi_3 / \pi_7 \approx 40$), we minimise the focal loss
\citep{lin2017focal} with label smoothing $\eta = 0.05$:
\begin{equation}
\mathcal{L}_{\mathrm{cls}}(\theta) \;=\;
\mathbb{E}_{(G,y)\sim\mathcal{D}_{\mathrm{train}}}\!
\Big[\, -\sum_{c \in \mathcal{C}} \alpha_c \,\big(1 - p_c\big)^{\gamma}\, \widetilde{q}_c \,\log p_c \Big],
\label{eq:focal}
\end{equation}
where $p_c = [f_\theta(G)]_c$,
$\alpha_c \propto 1/\pi_c$ are inverse-frequency weights,
$\gamma = 1$, and the smoothed target is
$\widetilde{q}_c = (1-\eta)\,\mathbbm{1}[c=y] + \tfrac{\eta}{|\mathcal{C}|-1}\mathbbm{1}[c\neq y]$.
At test time we report accuracy, macro-F1 \& macro-AUROC.

\section{Experiments}
\label{sec:experiments}
In this section, we run \& demonstrate the performance of our method, against structurally less expressive baselines (Section~\ref{subsec:results}), external baseline GNNs (Section~\ref{subsec:external-baselines}) \& across multiple tasks (Section~\ref{subsec:tasks}). Lastly, we provide extensive ablation of each component of our method.

\subsection{Dataset \& Metrics}

We use the ProteinShake \citep{kucera2023proteinshake},
15{,}603 PDB structures annotated with first-level EC numbers. Nodes are
residues \& edges connect C\textsubscript{$\alpha$} atoms within
$\varepsilon=8\,$\AA. We adopt ProteinShake's random split with a strict
sequence-similarity threshold $\theta=0.7$, yielding 14{,}042 train / 780
val / 781 test proteins. The dataset is severely imbalanced: EC 3 has 5{,}619
training samples, EC 7 only 139 (40$:\!1$ ratio). We report accuracy,
macro-F1, \& macro-AUROC; macro-F1 is our primary metric because it weights
rare classes equally. Experiments run on a single NVIDIA RTX 6000 Ada (48\,GB).

\subsection{Primary Results}
\label{subsec:results}

Table~\ref{tab:main} compares SoftBlobGIN against three baselines of
increasing structural expressivity (defined in Appendix~\ref{app:base}).
The hierarchy reveals a clear pattern: the largest performance jump comes
from adding ESM-2 features (Seq MLP $\to$ Residue MLP: $+0.199$ accuracy,
$+0.270$ macro-F1), not from adding graph structure (Residue MLP $\to$ GIN:
$+0.015$ accuracy, $+0.002$ macro-F1). This confirms that frozen PLM
embeddings are the dominant signal for graph-level EC classification.
Nonetheless, graph structure is not redundant. GIN achieves the highest
single-model accuracy (0.925), \& SoftBlobGIN achieves the highest
single-model macro-F1 (0.876), a metric that weights rare classes equally
\& is therefore more informative given the 40:1 class imbalance. The
5-seed ensemble stabilises minority-class variance
(Appendix~\ref{app:hparams}) \& reaches 0.928 accuracy \& 0.898
macro-F1. Crucially, SoftBlobGIN's competitive performance comes
\emph{with} built-in blob-level interpretability
(Section~\ref{sec:blobs}), whereas GIN's marginally higher accuracy
provides no such structural decomposition.

\begin{table}[ht]
\centering
\small
\begin{minipage}{0.5\linewidth}
\centering
\setlength{\tabcolsep}{5pt}
\begin{tabular}{@{}lcccrc@{}}
\toprule
Model & Acc & Mac.\ F1 & Mac.\ AUROC & Params & Time \\
\midrule
Seq MLP     & 0.711 & 0.597 & 0.912 & 140K & 54\,s \\
Residue MLP & 0.910 & 0.867 & 0.964 & 1.2M & 32\,s \\
GIN         & 0.925 & 0.869 & 0.969 & 1.4M & 15\,m \\
SoftBlobGIN & 0.912 & 0.876 & 0.951 & 1.1M & 27\,m \\
\midrule
\textbf{Ens.\ (5$\times$SoftBlobGIN)} & \textbf{0.928} & \textbf{0.898} & \textbf{0.955} & -- & -- \\
\bottomrule
\end{tabular}
\end{minipage}
\hfill
\begin{minipage}{0.27\linewidth}
\caption{\small EC classification on the ProteinShake test set (781 proteins, 7
classes). Baselines are defined in Appendix~\ref{app:base}. Macro-F1 is the
primary metric because it gives equal weight to each class despite the 40:1
imbalance. The ensemble averages softmax outputs across five random seeds.}
\label{tab:main}
\end{minipage}
\end{table}


\subsection{Comparison with external baselines}
\label{subsec:external-baselines}

We compare against two external structure-based baselines, GearNet
\citep{zhang2023gearnet} \& ProNet \citep{wang2023pronet}, as well as a
frozen ESM-2 mean-pooled linear probe, all evaluated on identical splits
(Table~\ref{tab:external}). The ESM-2 linear probe is a remarkably strong
baseline: $0.841$ accuracy \& $0.787$ macro-F1 with only 9K trainable
parameters, confirming that frozen PLM features already capture most of
the EC signal. GearNet \& ProNet, despite being substantially larger
(16.3M \& 0.73M parameters) \& designed for structure-based protein
learning, perform well below the linear probe on this split. Neither
architecture uses ESM-2 embeddings, relying instead on coordinate-derived
features alone; their low scores reflect the difficulty of learning
EC-discriminative representations from structure without the evolutionary
context that PLMs provide. The linear probe thus establishes the ceiling
of what ESM-2 can achieve without structural reasoning ($\sim\!0.84$
accuracy). SoftBlobGIN improves $+7.1$ accuracy points \& $+8.9$
macro-F1 points above this ceiling, \& this gain comes \emph{with} the
interpretability machinery of Section~\ref{sec:explanations}.

\begin{table}[ht]
\centering
\small
\begin{minipage}{0.45\linewidth}
\centering
\begin{tabular}{lccrr}
\toprule
Model                          & Accuracy & Macro F1 & Params & Time \\
\midrule
GearNet \citep{zhang2023gearnet}         & 0.539  & 0.190 & 16.3M & 63\,m \\
ProNet \citep{wang2023pronet}            & 0.558  & 0.439 & 0.73M & 21\,m \\
ESM-2 linear probe (mean-pool)           & 0.841  & 0.787 & 9.0K & 30\,m \\
\midrule
\textbf{SoftBlobGIN}                     & 0.912  & 0.876 & 1.1M & 27\,m \\
\textbf{Ensemble (5$\times$SoftBlobGIN)} & \textbf{0.928} & \textbf{0.898} & -- & -- \\
\bottomrule
\end{tabular}
\end{minipage}
\hfill
\begin{minipage}{0.25\linewidth}
\caption{\small External baseline comparison on ProteinShake EC using identical
splits. GearNet \& ProNet use coordinate-only features, whereas the linear
probe \& our models use frozen ESM-2 embeddings.}
\label{tab:external}
\end{minipage}
\end{table}

\subsection{ProteinShake benchmark sweep}
\label{subsec:tasks}

We apply the same architecture (with task-specific output heads: linear
classifier, multi-label sigmoid, scalar regression, node binary
classifier, Siamese, bilinear PPI) to all ProteinShake tasks
(Table~\ref{tab:proteinshake}). The same SoftBlobGIN backbone is
competitive across heterogeneous task types. Binding-site detection at
the node level shows the largest delta over the linear probe ($+0.086$
AUROC, MCC $0.783$), confirming the $+9.8$-point gap reported in
Section~\ref{sec:esm-vs-gnn} \& reflecting the strong inductive bias
of message passing for residue-level predictions. Pairwise
StructureSimilarity also shows a large gain ($+0.193$ Spearman), as
expected for a task that intrinsically depends on 3D structure.
ProteinFamily \& StructuralClass are nearly saturated by the linear
probe, indicating that ESM-2 already encodes most of the
fold-classification signal \& graph structure adds little beyond it.
LigandAffinity remains the weakest area in absolute terms, possibly
reflecting that affinity depends on details beyond the residue contact
graph.

\begin{table}[ht]
\centering
\small
\setlength{\tabcolsep}{4pt}
\begin{tabular}{llccccc}
\toprule
Task & Metric
     & GearNet & ProNet & ESM-2 probe & SoftBlobGIN & $\Delta_{\text{probe}}$ \\
\midrule
EnzymeClass        & Accuracy       & 0.539 & 0.558 & \underline{0.841} & \textbf{0.912} & $+$0.071 \\
GeneOntology       & $F_{max}$      & 0.560 & 0.573 & \underline{0.721} & \textbf{0.733} & $+$0.012 \\
ProteinFamily      & Accuracy       & 0.014 & 0.325 & \underline{0.834} & \textbf{0.839} & $+$0.005 \\
StructuralClass    & Accuracy       & 0.012 & 0.109 & \textbf{0.698}    & \underline{0.696} & $-$0.002 \\
LigandAffinity     & R$^2$          & -   & 0.372 & 0.372             & \textbf{0.422} & $+$0.050 \\
BindingSite        & AUROC          & \underline{0.954} & 0.903 & 0.883 & \textbf{0.969} & $+$0.086 \\
Struct.\ Similarity & Spearman      & 0.534 & \underline{0.645} & 0.523 & \textbf{0.716} & $+$0.193 \\
\bottomrule
\end{tabular}

\vspace{1mm}

\caption{\small ProteinShake benchmark (random split). Each cell reports the task’s official primary metric. Best results are in \textbf{bold}; second-best are \underline{underlined}. `-` denotes failed or inapplicable methods. GearNet \& ProNet use only coordinate features, while the linear probe \& SoftBlobGIN use frozen ESM-2 embeddings.}
\label{tab:proteinshake}
\end{table}

\subsection{Ablation Studies}

To isolate the contribution of each feature group, we retrain SoftBlobGIN
with progressively richer input representations
(Table~\ref{tab:abl-feats}). One-hot amino-acid identity alone yields
0.752 accuracy \& 0.624 macro-F1. Appending physicochemical properties,
SASA/RSA, \& edge features slightly \emph{decreases} raw accuracy
(0.722) while improving macro-F1 (0.653), reflecting better calibration
on rare classes at the expense of majority-class accuracy. The single
largest jump occurs when frozen ESM-2 embeddings are added: macro-F1
rises from 0.653 to 0.853, accounting for approximately 85\% of the
total improvement. The full feature set provides the highest AUROC
(0.978), though macro-F1 dips marginally to 0.846, suggesting mild
redundancy between ESM-2 \& handcrafted features that the model
resolves differently across metrics. Additional hyperparameter sweeps
over contact radius ($\varepsilon \in \{4, 8, 12\}\,$\AA, optimal at
$8$\,\AA) \& blob count ($K \in \{3, 5, 8, 12\}$, optimal at
$K{=}8$) are reported in Appendix~\ref{app:hparams}. The 5-seed
ensemble adds $+0.040$ macro-F1 over the mean individual model.

\begin{table}[ht]
\centering
\small
\begin{tabular}{lcccc}
\toprule
Feature set                       & Dim   & Accuracy & Macro F1 & AUROC \\
\midrule
One-hot only                      & 20    & 0.752    & 0.624    & 0.890 \\
$+$ Physico $+$ SASA $+$ Edge     & 38    & 0.722    & 0.653    & 0.940 \\
$+$ ESM-2 (1280)                  & 1318  & 0.909    & 0.853    & 0.966 \\
Full (all features)               & 1318  & 0.913    & 0.846    & 0.978 \\
\bottomrule
\end{tabular}
\vspace{1mm}
\caption{\small Feature-set ablation on SoftBlobGIN (EC test set). Each row
cumulatively adds one feature group to the row above. Dim is the resulting
node-feature dimensionality. The ESM-2 row produces the single largest
gain ($+0.200$ macro-F1), confirming that frozen PLM embeddings are the
dominant contributor to classification performance.}
\label{tab:abl-feats}
\end{table}

\section{Interpretable Structural Explanations}
\label{sec:explanations}

\subsection{When does structure help? Binding-site as a clean test}
\label{sec:esm-vs-gnn}

The interpretability claim is only meaningful if a structurally aware GNN
\emph{adds} something ESM-2 alone does not provide. We test this on
ProteinShake binding-site detection ($n{=}465$ test proteins with
per-residue ground-truth labels), comparing three approaches that all
consume the same frozen ESM-2 features (Table~\ref{tab:probe}):
unsupervised ESM-2 attention; a per-residue linear probe on ESM-2; \&
the GIN backbone with message-passing over the contact graph. Blob
pooling is bypassed for this node-level task (Section~\ref{sec:method}); the
result reflects the GIN backbone alone.

\begin{table}[ht]
\centering
\small
\begin{minipage}{0.56\linewidth}
\centering
\begin{tabular}{lcc}
\toprule
Method & Pred AUROC & Top-10\% Prec. \\
\midrule
ESM-2 attention (unsupervised)
    & $0.634 \pm 0.095$
    & $0.331 \pm 0.187$ \\
Linear probe on ESM-2
    & $0.885 \pm 0.119$
    & $0.758 \pm 0.240$ \\
\textbf{GIN backbone (ESM-2 + graph)}
    & $\mathbf{0.983 \pm 0.040}$
    & $\mathbf{0.931 \pm 0.153}$ \\
\bottomrule
\end{tabular}
\end{minipage}
\hfill
\begin{minipage}{0.28\linewidth}
\caption{\small Binding-site prediction quality on $n{=}465$ test proteins. All
methods use the same frozen ESM-2 features \& differ only in how those
representations are aggregated.}
\label{tab:probe}
\end{minipage}
\end{table}

The 9.8-point AUROC gap between the GIN \& the linear probe ($0.983$ vs
$0.885$) is the central evidence that graph structure adds binding-site
information ESM-2 alone cannot recover. Both methods use identical
features; the delta is wholly attributable to message-passing over the
contact graph. ESM-2 attention is much weaker still ($0.634$); its high
entropy ($5.12$) \& low sequence bias ($0.068$) confirm that attention
is broadly diffuse rather than anchored to spatially specific binding
pockets. We use GNNExplainer \& Integrated Gradients rather than vanilla
gradient saliency, which is uninformative on the trained GNN due to
saturation at high-confidence residues (Appendix~\ref{app:saliency}).

\paragraph{When does structure not help?} As an out-of-distribution
probe, we evaluated SoftBlobGIN on DeepLoc-2.1 subcellular localisation
\citep{thumuluri2022deeploc}. ESM-only reaches $F_{max} = 0.688$;
SoftBlobGIN reaches $0.682$, indicating that adding the GNN does not
improve over ESM-only. Localisation depends on signal peptides \&
transmembrane segments rather than residue contacts, consistent with the
dichotomy that the structural companion adds value when the prediction
is structurally mediated \& contributes little when it is not.

\subsection{Biological co-referencing of GNNExplainer outputs}
\label{subsec:coref}

\begin{figure*}[ht]
    \centering
    \begin{minipage}{0.32\textwidth}
        \centering
        \includegraphics[width=\linewidth]{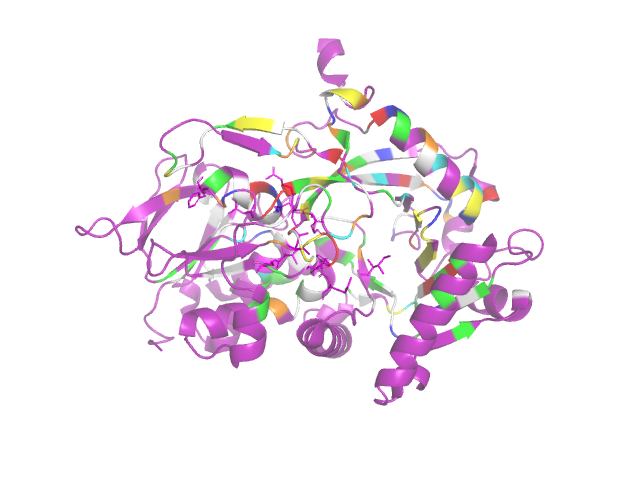}
        
        \vspace{2mm}
        \footnotesize
        \textbf{4RSL} \\
        EC 3: 
        Blob 2 captures catalytic pocket
    \end{minipage}
    \hfill
    \begin{minipage}{0.32\textwidth}
        \centering
        \includegraphics[width=\linewidth]{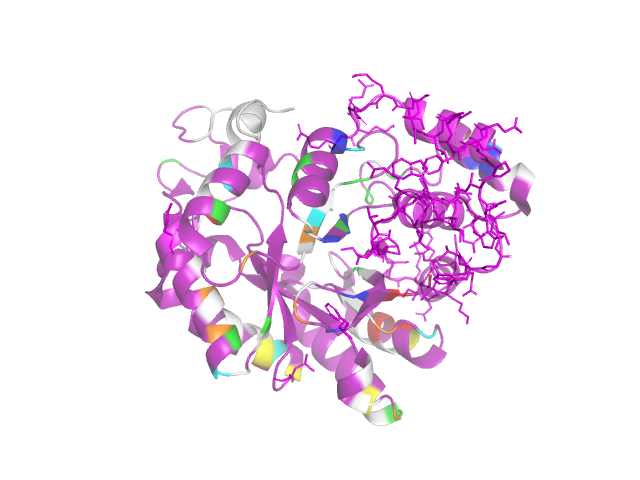}
        
        \vspace{2mm}
        \footnotesize
        \textbf{2VDG} \\
        EC 6: 
        Blob 4 localizes active-site residues
    \end{minipage}
    \hfill
    \begin{minipage}{0.32\textwidth}
        \centering
        \includegraphics[width=\linewidth]{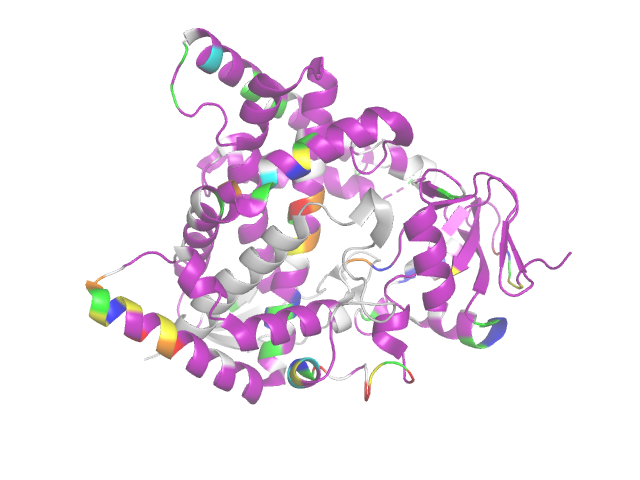}
        
        \vspace{2mm}
        \footnotesize
        \textbf{6A18} \\
        EC 7: 
        Blob 1 spans transport core
    \end{minipage}
    
    \caption{\small 
    Qualitative 3D case studies of learned SoftBlobGIN substructures.
    Proteins are rendered with residues colored by blob assignment, while
    annotated active-site residues are shown as magenta sticks. Across
    single-domain, multi-domain, \& translocase examples, one dominant
    learned blob consistently overlaps functional active region,
    supporting biological interpretability of learned partitions.
    }
    \label{fig:blob_case_studies}
\end{figure*}

Predictive faithfulness is necessary but insufficient: explanations
must also align with known enzyme biochemistry. We evaluate the four
biological-faithfulness criteria from Section~\ref{sec:problem} on the
top-20\% residues per protein ranked by GNNExplainer importance
(Table~\ref{tab:bio-aa}). \textbf{(B1)} Top residues are enriched for
known catalytic amino acids in the EC classes where these residues are
catalytically established: Cys/His for oxidoreductases (EC 1),
Ser/His/Asp for hydrolases (EC 3), \& charged residues for translocases
(EC 7). \textbf{(B2)} The empirical SASA gap is negative for all 7 EC
classes; important residues are consistently more buried than
unimportant ones, the structural signature of active-site pockets.
\textbf{(B3)} The spatial $z$-score
$\mathbb{E}_G[Z_{\mathrm{spatial}}(G)] < 0$ across all classes, with
important residues $5$--$15\%$ more compact in 3D than random subsets.
\textbf{(B4)} Among edges with mask values $\geq 0.5$, the
C\textsubscript{$\alpha$}--C\textsubscript{$\alpha$} distance distribution
peaks in $[6, 10]\,$\AA, the catalytic-triad regime (Ser--His
${\sim}6$\,\AA, His--Asp ${\sim}7$\,\AA{} in serine proteases).

\begin{table}[ht]
\centering
\small
\setlength{\tabcolsep}{4pt}
\renewcommand{\arraystretch}{1.15}

\begin{tabularx}{\columnwidth}{
>{\raggedright\arraybackslash}p{0.18\columnwidth}
>{\raggedright\arraybackslash}p{0.37\columnwidth}
>{\raggedright\arraybackslash}X
}
\toprule
\textbf{EC class} &
\textbf{GNNExplainer highlights} &
\textbf{Known biological interpretation} \\
\midrule

\textbf{EC 1} \newline Oxidoreductase
&
Physicochemical feature group ranked highest; Cys \& His enriched
&
Redox-active cofactors commonly coordinated through Cys/His ligation
\\

\textbf{EC 3} \newline Hydrolase
&
SASA ranked highest; Ser, His, \& Asp enriched
&
Consistent with the canonical Ser--His--Asp catalytic triad
\\

\textbf{EC 5} \newline Isomerase
&
SASA \& degree features dominant; His \& Lys enriched
&
Reflects catalytic hub residues \& Lys-mediated Schiff-base chemistry
\\

\textbf{EC 6} \newline Ligase
&
Physicochemical feature group ranked highest; densest masks ($75.3\%$)
&
Matches multi-domain ATP-binding architectures \& buried active sites
\\

\textbf{EC 7} \newline Translocase
&
SASA ranked highest; Asp, Glu, Lys, \& Arg enriched; sparsest masks
&
Consistent with transmembrane topology \& charged transport residues
\\

\bottomrule
\end{tabularx}

\vspace{1mm}
\caption{
\small Top-ranked feature groups \& enriched amino acids identified by
GNNExplainer across EC classes. Explanations recover class-specific
biochemical motifs \& structural signatures.
}
\label{tab:bio-aa}
\end{table}

\subsection{Learned blobs map to functional substructures}
\label{sec:blobs}

SoftBlobGIN's blob assignments provide an explanation by construction,
without any post-hoc optimisation. Across all test proteins, the model
spontaneously decomposes each structure into a large, solvent-exposed
structural-core blob (102 to 335 residues, mean SASA $0.28$ to $0.36$)
\& several small, buried functional-site blobs (8 to 40 residues, mean
SASA $0.10$ to $0.17$). This core-versus-functional separation is
consistent across all 7 EC classes, with functional blobs
\textbf{2 to 2.6$\times$ more buried} than the core blob, despite the
model receiving no active-site supervision during training
(per-blob statistics in Appendix~\ref{app:blob-stats}).

To test whether this decomposition captures functionally relevant
substructures, we analyse blob importance scores for 59 EC test proteins
with active-site annotations obtained via the PDB$\,\to\,$UniProt
features pipeline. We define the importance $\pi_t$ of blob $t$ as the
fraction of readout dimensions in which blob $t$ supplies the argmax in
$z_{\mathrm{blob}} = \max_k b_k$ (Eq.~\ref{eq:blob-readout}), analogous
to BioBlobs' attention weight. The blob containing the most annotated
active-site residues (the \emph{active blob}) carries higher importance
than competing blobs in \textbf{62.7\% of proteins}, with mean
importance \textbf{$0.209$ vs.\ $0.113$} for other blobs
(\textbf{1.85$\times$ enrichment}). On average, the active blob ranks
\textbf{$2.97/8$} (top \textbf{$37\%$}), and the correlation between
$\pi_t$ rank and active-site enrichment is statistically significant
(\textbf{Spearman $\rho = 0.339$, $p = 0.009$}). The effect is moderate because EC classification depends on both
active-site chemistry \& global fold topology, but the signal is
reproducible across the test set. Together with the burial separation
above, these analyses indicate that learned blobs capture
\textbf{biologically meaningful substructures} rather than arbitrary
partitions. The partial overlap with Pfam/CATH domain boundaries
(Jaccard \textbf{$0.380$}, decaying with domain count;
Appendix~\ref{app:domain}) further suggests that blobs are optimised
for the classification objective rather than recapitulating known
domain segmentations.

\section{Discussion \& Limitations}
\label{sec:discussion}

\paragraph{ESM-2 dominates EC classification, but not binding-site
prediction.} A simple MLP on mean-pooled ESM-2 nearly matches the full GIN
on EC classification (0.867 vs 0.869 macro-F1). For first-level EC
classification, ESM-2 has likely seen enough evolutionary diversity during
pretraining to implicitly encode the relevant motifs. The picture is
\emph{very different} on binding-site detection
(Section~\ref{sec:esm-vs-gnn}): ESM-2 attention reaches only AUROC $0.634$,
an ESM-2 linear probe reaches $0.885$, \& SoftBlobGIN reaches $0.983$.
Using the same features, graph message passing alone adds $+9.8$ absolute
AUROC. This is the cleanest evidence in the paper that structural
reasoning is not redundant: it adds binding-site-relevant information that
ESM-2 alone cannot recover \&, together with the
biological-faithfulness analyses, makes the prediction auditable.


\paragraph{Interpretability is more than fidelity.} Our explanations have
imperfect Fid$^-$ (the model is too redundant to be ``broken'' by edge
removal) but excellent biological alignment (catalytic residues, burial,
spatial clustering, contact geometry). We argue that biological
co-referencing is the more meaningful test for protein-function explanation,
\& we encourage the community to evaluate explainers against domain-derived
priors rather than only against the explainer's own optimisation objective.

\paragraph{Limitations.} The model exhibits substantial redundancy,
driving Fid$^-$ close to zero because multiple edge subsets can support
the same prediction. Gradient saturation at high-confidence residues
also limits vanilla input$\,\times\,$gradient saliency
(Appendix~\ref{app:saliency}), motivating our use of GNNExplainer \&
Integrated Gradients \citep{yuan2022explainability}. Biological validation is based on general biochemical priors rather than
residue-level catalytic annotations; incorporating curated supervision
such as the Catalytic Site Atlas would enable finer-grained evaluation.
Learned blobs capture features useful for EC classification but do not
explicitly align with Pfam/CATH domains (Jaccard $0.380$;
Appendix~\ref{app:domain}). SoftBlobGIN also tends to allocate
$50$--$60\%$ of residues to a dominant core blob, suggesting room for
improved partition balance.

\section{Conclusion}
\label{sec:conclusion}

We presented \textit{\textbf{SoftBlobGIN}}, a lightweight graph neural network that learns differentiable protein
substructure partitions. Across the
ProteinShake benchmark, the framework achieves strong classification
performance while producing structurally grounded explanations:
\textit{GNNExplainer} recovers catalytic residues, active-site burial patterns,
\& contact geometry consistent with known enzyme biochemistry, \&
learned blobs spontaneously separate functional sites from structural
scaffold without active-site supervision. The framework is plug-\&-play (no PLM retraining), adds only $\sim$1.1M parameters, \& generalises across ProteinShake tasks. We
position this not as a replacement for protein language models but as an
\emph{interpretable structural companion} that makes their predictions
auditable for downstream scientific \& clinical use.


\newpage
\bibliographystyle{plainnat}
\bibliography{refs}
\newpage
\appendix
\section{Algorithmic Details}
\label{app:alg}
\begin{algorithm}[ht]
\caption{SoftBlobGIN forward pass}
\label{alg:sbgin}
\begin{algorithmic}[1]
\Require Graph $G=(V,E,\mathbf{X},\mathbf{E}_{\mathrm{attr}})$, temperature $\tau$
\State $H^{(0)} \gets \mathrm{Linear}(\mathbf{X})$
\For{$\ell = 1, \dots, L$}
   \State $H^{(\ell)} \gets \mathrm{GINEConv}(H^{(\ell-1)}, E, \mathbf{E}_{\mathrm{attr}})$
   \State $H^{(\ell)} \gets \mathrm{Drop}(\mathrm{ReLU}(\mathrm{BN}(H^{(\ell)})))$
\EndFor
\State $L \gets \mathrm{BlobHead}(H^{(L)})$ ; \quad
       $A \gets \mathrm{GumbelSoftmax}(L, \tau)$
\For{$k = 1,\dots,K$}
   \State $b_k \gets \sum_i A_{ik} h_i^{(L)} / (\sum_i A_{ik}+\epsilon)$;
          \quad $b_k \gets \mathrm{LN}(\mathrm{MLP}(b_k))$
\EndFor
\State $z_{\mathrm{blob}} \gets \max_k b_k$;
       \quad $z_{\mathrm{global}} \gets \tfrac{1}{N}\sum_i h_i^{(L)}$
\State \Return $\mathrm{Classifier}([z_{\mathrm{blob}} \,\|\, z_{\mathrm{global}}])$
\end{algorithmic}
\end{algorithm}

\section{Node Features}
\label{app:feat}

Each residue $i$ carries a
$d$-dimensional feature vector $x_i \in \mathbb{R}^d$ formed by
concatenating six blocks:
\begin{equation}
x_i \;=\; \big[\, \phi^{\mathrm{aa}}(s_i) \,\|\, \phi^{\mathrm{phys}}(s_i) \,\|\, \phi^{\mathrm{sasa}}(i) \,\|\,
\phi^{\mathrm{esm}}(S)_i \,\|\, \phi^{\mathrm{deg}}(i) \,\|\, \phi^{\mathrm{pos}}(i, N) \,\big],
\label{eq:node-feat}
\end{equation}
with dimensions $20 + 10 + 2 + 1280 + 1 + 5 = 1318$. Specifically:
\begin{itemize}[leftmargin=1.3em,itemsep=1pt,topsep=2pt]
\item $\phi^{\mathrm{aa}}(s) \in \{0,1\}^{20}$ is the one-hot encoding;
\item $\phi^{\mathrm{phys}}(s) \in [0,1]^{10}$ is the min--max normalised
      vector of (Kyte--Doolittle hydrophobicity, charge, MW, vdW volume,
      Grantham polarity, Vihinen flexibility, accessibility, helix-/sheet-/
      turn-propensity);
\item $\phi^{\mathrm{sasa}}(i) = (\widetilde{\mathrm{SASA}}_i,
      \mathrm{RSA}_i) \in [0,1]^2$;
\item $\phi^{\mathrm{esm}}: \Sigma^* \to \mathbb{R}^{N\times 1280}$ is the
      ESM-2 (650M, layer 33) per-residue extractor
      \citep{lin2023evolutionary}, treated as a frozen black box;
\item $\phi^{\mathrm{deg}}(i) = \deg(i)/\max_{j} \deg(j) \in [0,1]$;
\item $\phi^{\mathrm{pos}}(i,N) = \big(\tfrac{i}{N},
      \sin\tfrac{i\pi}{N}, \cos\tfrac{i\pi}{N},
      \sin\tfrac{2i\pi}{N}, \cos\tfrac{2i\pi}{N}\big)$.
\end{itemize}
We write $\mathbf{X} \in \mathbb{R}^{N \times d}$ for the stack of $x_i$.

\section{Metrics}
\label{app:met}

We measure explanation quality via four scalar functionals on
$\mathcal{D}_{\mathrm{test}}$. Let $\mathcal{T}_s(M)$ denote the binary mask
that retains only the top $\lceil s|E|\rceil$ edges of $M$, \& let
$\mathbf{1}\!-\!\mathcal{T}_s(M)$ be its complement.

\paragraph{(a) Sparsity.}
$\mathrm{Sp}(M) = |E|^{-1}\sum_e \mathbbm{1}[M_e < 0.5].$

\paragraph{(b) Sufficiency (Fidelity\textsuperscript{$+$}).} Predictive
agreement when keeping only the top edges:
\begin{equation}
\mathrm{Fid}^{+}(s) \;=\; \mathbb{E}_{(G,y)}\!\Big[\, \mathbbm{1}\big[\, \arg\max f_\theta(G_{\mathcal{T}_s(M)}) = \arg\max f_\theta(G) \,\big] \,\Big].
\label{eq:fidplus}
\end{equation}

\paragraph{(c) Necessity (Fidelity\textsuperscript{$-$}).} Predictive
breakage when removing the top edges:
\begin{equation}
\mathrm{Fid}^{-}(s) \;=\; 1 - \mathbb{E}_{(G,y)}\!\Big[\, \mathbbm{1}\big[\, \arg\max f_\theta(G_{\mathbf{1}-\mathcal{T}_s(M)}) = \arg\max f_\theta(G) \,\big] \,\Big].
\label{eq:fidminus}
\end{equation}

\paragraph{(d) Class stability.} Mean intra-class cosine similarity of
feature masks:
\begin{equation}
\mathrm{Stab}(c) \;=\; \frac{2}{|\mathcal{S}_c|(|\mathcal{S}_c|-1)} \sum_{\substack{(F^{(a)}, F^{(b)}) \in \mathcal{S}_c \\ a < b}} \frac{\langle F^{(a)}, F^{(b)}\rangle}{\|F^{(a)}\|\,\|F^{(b)}\|},
\end{equation}
where $\mathcal{S}_c = \{F^{(n)} : y^{(n)} = c\}$.

\section{Baseline Methods}
\label{app:base}
We evaluate four baseline GNNs of increasing structural expressivity
(Table~\ref{tab:models}). All four share the feature pipeline of
Eq.~\ref{eq:node-feat} so that differences in
performance reflect differences in topological inductive bias.
\begin{table}[ht]
\centering
\small
\begin{tabular}{lll}
\toprule
Model & Input & Architecture \\
\midrule
Seq MLP        & AA composition (20-d)             & 3-layer MLP \\
Residue MLP    & Mean-pool of 1318-d features      & 3-layer MLP \\
GIN            & Graph + 1318-d nodes + 18-d edges & GINEConv $\times$ 4 + JK + dual pool \\
SoftBlobGIN    & Graph + features                  & GIN backbone + Gumbel blob pool \\
\bottomrule
\end{tabular}

\vspace{1mm}

\caption{Model hierarchy by structural expressivity.}
\label{tab:models}
\end{table}
\paragraph{Seq MLP \& Residue MLP.} The first two baselines deliberately
\emph{discard} graph structure to isolate the contribution of the contact
graph. Seq MLP uses only the bag of amino acids vector
$\bar{x}^{\mathrm{aa}} = \tfrac{1}{N}\sum_{i} \phi^{\mathrm{aa}}(s_i)$ as
input, while Residue MLP uses the full mean-pooled feature
$\bar{x} = \tfrac{1}{N}\sum_{i} x_i \in \mathbb{R}^{1318}$. Both end in a
3-layer MLP with LayerNorm, GELU \& dropout 0.3.

\paragraph{GIN.} We instantiate $g_\theta$ with $L = 4$ GINEConv \citep{hu2019strategies} layers of width $\hbar = 256$, BatchNorm \& ReLU
between layers, JumpingKnowledge concatenation \citep{xu2018jumping}
($\hbar L = 1024$ per node), \& a dual mean-max
readout to a 2048-dim graph embedding. A two-layer MLP head with
BatchNorm produces the final logits. Total: $\sim$1.4M parameters.

\paragraph{SoftBlobGIN.} The final architecture replaces JumpingKnowledge
\& the dual readout with the differentiable soft-blob pooling defined in
Eq.~(\ref{eq:gumbel}) through (\ref{eq:readout}). We use $K = 8$
blobs \& anneal $\tau$ linearly from $1.0$ to $0.1$ across training
epochs. The blob refinement $g_\psi$ is a 2-layer MLP with LayerNorm. This
substitutes BioBlobs' GVP encoder + VQ codebook with a single MLP head, at
$\sim$1.1M total parameters. Algorithm~\ref{alg:sbgin} gives the full
forward pass.

\section{Hyperparameter Sweeps}
\label{app:hparams}
We searched over contact radius
$\varepsilon \in \{4, 8, 12\}\,$\AA{} \& blob count
$K \in \{3, 5, 8, 12\}$ on the EC validation split. Performance peaked
at $\varepsilon = 8\,$\AA{} (the standard choice for residue contact
graphs) \& $K = 8$ blobs, consistent with the intuition that enzymes
have a small number of distinct functional substructures (active site,
cofactor pocket, substrate channel, scaffold). At $K = 3$ the model
under-segments \& core blobs absorb functional residues; at $K = 12$
many blobs are empty after Gumbel annealing. Performance is broadly
stable in the $K \in [5, 8]$ range.

\begin{table}[ht]
\centering
\small
\begin{minipage}[t]{0.60\textwidth}
\centering
\begin{tabular}{lccc}
\toprule
Seed              & Accuracy & Macro F1 & AUROC \\
\midrule
42                & 0.914 & 0.890 & 0.943 \\
123               & 0.909 & 0.871 & 0.939 \\
456               & 0.890 & 0.824 & 0.948 \\
789               & 0.912 & 0.888 & 0.949 \\
1024              & 0.898 & 0.815 & 0.931 \\
\midrule
Mean individual   & 0.905 & 0.858 & 0.942 \\
\textbf{Ensemble} & \textbf{0.928} & \textbf{0.898} & \textbf{0.955} \\
Ensemble gain     & $+0.024$ & $+0.040$ & $+0.013$ \\
\bottomrule
\end{tabular}
\end{minipage}%
\hfill
\begin{minipage}[t]{0.35\textwidth}
\vspace{0pt}
\caption{5-seed SoftBlobGIN ensemble on ProteinShake EC.}
\label{tab:ensemble}
\end{minipage}
\end{table}

\section{Explainability Analysis}
\label{app:saliency}

We report quantitative explanation metrics for SoftBlobGIN \& Integrated Gradients (IG),
evaluating sparsity, stability, \& faithfulness of learned explanations on the ProteinShake EC benchmark.
Vanilla gradient saliency was excluded due to gradient saturation, producing noisy, poorly localized scores.

\paragraph{Aggregate statistics.}
Table~\ref{tab:explanation_summary} summarizes explanation quality for SoftBlobGIN.
Explanations are highly sparse (mean $= 0.814$), retaining only a small fraction of residues with high attribution.
Masked confidence of $0.736$ indicates that these residues alone preserve most predictive signal.
Stability is consistently high across all seven EC classes ($0.863$--$0.911$), confirming reproducible explanations within each functional family.

\begin{table}[ht]
\centering

\begin{minipage}{0.55\linewidth}
\centering
\small
\setlength{\tabcolsep}{12pt}
\renewcommand{\arraystretch}{1.15}
\begin{tabular}{lc}
\toprule
Metric & Value \\
\midrule
Mean sparsity                        & 0.814 \\
Mean masked confidence\,$\uparrow$   & 0.736 \\
Mean unfaithfulness\,$\downarrow$    & 0.437 \\
Characterization score               & 0.116 \\
EC stability range\,$\uparrow$       & 0.863--0.911 \\
\bottomrule
\end{tabular}
\end{minipage}
\hfill
\begin{minipage}{0.38\linewidth}
\small
\captionof{table}{Aggregate explanation statistics for SoftBlobGIN ($\uparrow$ higher is better; $\downarrow$ lower is better).}
\label{tab:explanation_summary}
\end{minipage}

\end{table}

\paragraph{Faithfulness under progressive masking.}
Table~\ref{tab:fidelity_compare} compares IG \& SoftBlobGIN via Fidelity$+$ (prediction preserved after retaining top residues) \& Fidelity$-$ (prediction degraded after removing them).
Both methods achieve Fidelity$+\!=\!1.000$ at $\geq\!20\%$ sparsity, confirming that top-ranked residues suffice to recover the original prediction.
Low Fidelity$-$ across all levels reflects mild redundancy in residue importance, consistent across both methods.

\begin{table}[ht]
\centering

\begin{minipage}{0.58\linewidth}
\centering
\small
\setlength{\tabcolsep}{8pt}
\renewcommand{\arraystretch}{1.12}
\begin{tabular}{lS[table-format=1.2]S[table-format=1.3]S[table-format=1.3]}
\toprule
Method & {Sparsity} & {Fidelity$+$ $\uparrow$} & {Fidelity$-$ $\uparrow$} \\
\midrule
\multirow{6}{*}{Integrated Gradients}
 & 0.05 & 0.978 & 0.000 \\
 & 0.10 & 0.989 & 0.011 \\
 & 0.20 & 1.000 & 0.011 \\
 & 0.30 & 1.000 & 0.022 \\
 & 0.50 & 1.000 & 0.022 \\
 & 0.70 & 1.000 & 0.022 \\
\addlinespace[3pt]
\multirow{6}{*}{SoftBlobGIN}
 & 0.05 & 0.989 & 0.000 \\
 & 0.10 & 0.978 & 0.000 \\
 & 0.20 & 1.000 & 0.011 \\
 & 0.30 & 1.000 & 0.011 \\
 & 0.50 & 1.000 & 0.022 \\
 & 0.70 & 1.000 & 0.022 \\
\bottomrule
\end{tabular}
\end{minipage}
\hfill
\begin{minipage}{0.36\linewidth}
\small
\captionof{table}{Faithfulness comparison under progressive sparsification. Fidelity$+$ $\uparrow$ measures retention of top residues; Fidelity$-$ $\uparrow$ measures degradation upon their removal.}
\label{tab:fidelity_compare}
\end{minipage}

\end{table}

\section{Blob structural statistics \& functional analysis}
\label{app:blob-stats}

SoftBlobGIN provides explanations by construction through learned blob
assignments, without requiring post-hoc optimization or gradient-based
attribution. To evaluate whether these assignments correspond to
biologically meaningful structural units, we analyze solvent
accessibility, active-site enrichment, amino acid composition, spatial
coherence, \& agreement with post-hoc explainers.

\paragraph{Structural decomposition of learned blobs.}
Across all test proteins, SoftBlobGIN consistently partitions each
protein into one large solvent-exposed structural blob \& several
smaller buried blobs. The dominant structural blob spans
$102$--$335$ residues with mean normalized solvent accessibility
(SASA) $0.28$--$0.36$, whereas smaller blobs contain $8$--$40$
residues with mean SASA $0.10$--$0.17$.

Functional blobs are therefore approximately $2$--$2.6\times$ more
buried than the structural core blob, despite the model receiving no
active-site supervision during training.

\begin{table}[ht]
\centering
\small
\setlength{\tabcolsep}{9pt}
\renewcommand{\arraystretch}{1.15}
\begin{tabular}{lccc}
\toprule
Blob type & Residues / blob & Mean normalized SASA & Interpretation \\
\midrule
Structural core blob & $102$--$335$ & $0.28$--$0.36$ & Surface-exposed scaffold \\
Functional-site blobs & $8$--$40$ & $0.10$--$0.17$ & Buried catalytic regions \\
\bottomrule
\end{tabular}

\vspace{1mm}

\caption{Aggregate structural statistics of learned blob assignments.}
\label{tab:blob_summary}
\end{table}

\begin{figure}[ht]
\centering
\includegraphics[width=0.95\columnwidth]{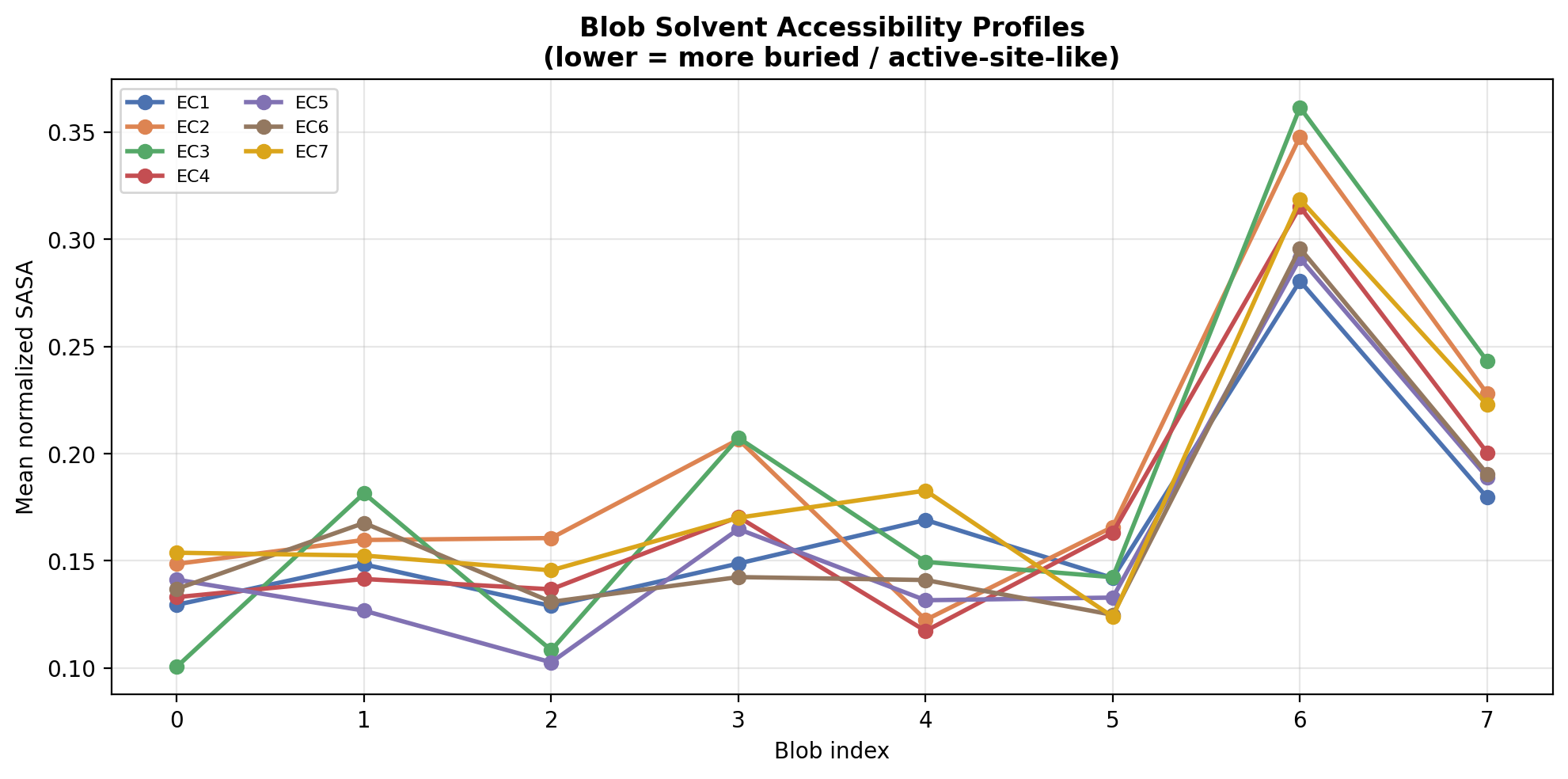}
\caption{Mean normalized solvent accessibility (SASA) of learned blobs
across EC classes. Lower values indicate more buried,
active-site-like regions.}
\label{fig:blob_sasa_profiles}
\end{figure}

Figure~\ref{fig:blob_sasa_profiles} shows solvent accessibility
profiles across EC classes. All classes exhibit a consistent pattern
with one dominant surface-exposed blob \& multiple buried blobs.

\paragraph{Active-site blob enrichment.}
To test whether learned blobs capture functionally relevant
substructures, we analyze proteins with active-site annotations
obtained through the PDB-to-UniProt features pipeline.

Blob importance is defined as the fraction of readout dimensions for
which blob $t$ provides the argmax in the pooled representation
$z_{\mathrm{blob}} = \max_k b_k$.

Across proteins with active-site annotations, blobs containing
annotated active-site residues carry $1.85\times$ higher importance
than other blobs on average, rank in the top 3 of 8, \& show
significant correlation between importance rank \& active-site
enrichment ($\rho = 0.339$, $p = 0.009$).

\begin{figure*}[ht]
\centering
\includegraphics[width=\textwidth]{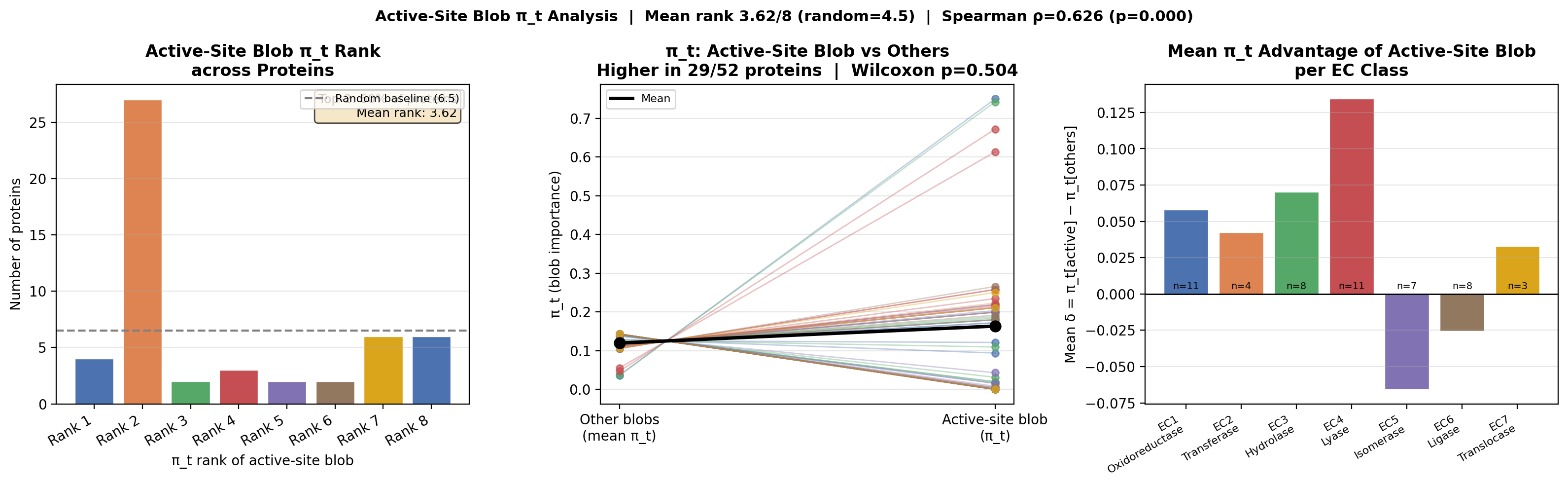}
\caption{Blob importance analysis for proteins with active-site
annotations. Left: rank distribution of active-site blobs. Middle:
comparison of active-site blob importance versus other blobs. Right:
mean importance advantage by EC class.}
\label{fig:active_site_blob_analysis}
\end{figure*}

\begin{figure}[ht]
\centering
\includegraphics[width=\columnwidth]{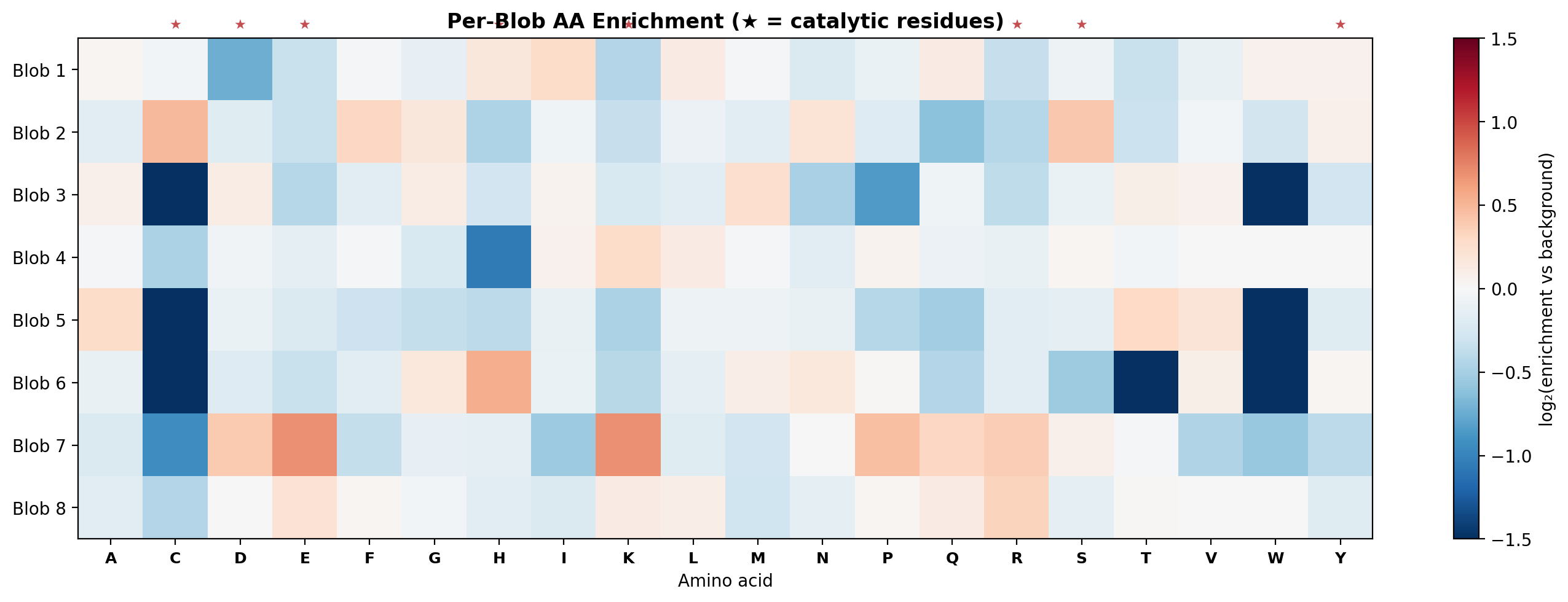}
\caption{Per-blob amino acid enrichment relative to background amino
acid frequencies. Stars indicate residues commonly associated with
catalytic activity.}
\label{fig:blob_aa_enrichment}
\end{figure}

\paragraph{Amino acid enrichment of learned blobs.}
To assess whether blobs capture chemically meaningful residue
composition, we compute amino acid enrichment relative to background
frequency for each blob. Figure~\ref{fig:blob_aa_enrichment} shows that several compact blobs
exhibit distinctive residue preferences, including enrichment of
residue types commonly associated with catalytic activity.

\begin{figure}[ht]
\centering
\includegraphics[width=0.9\columnwidth]{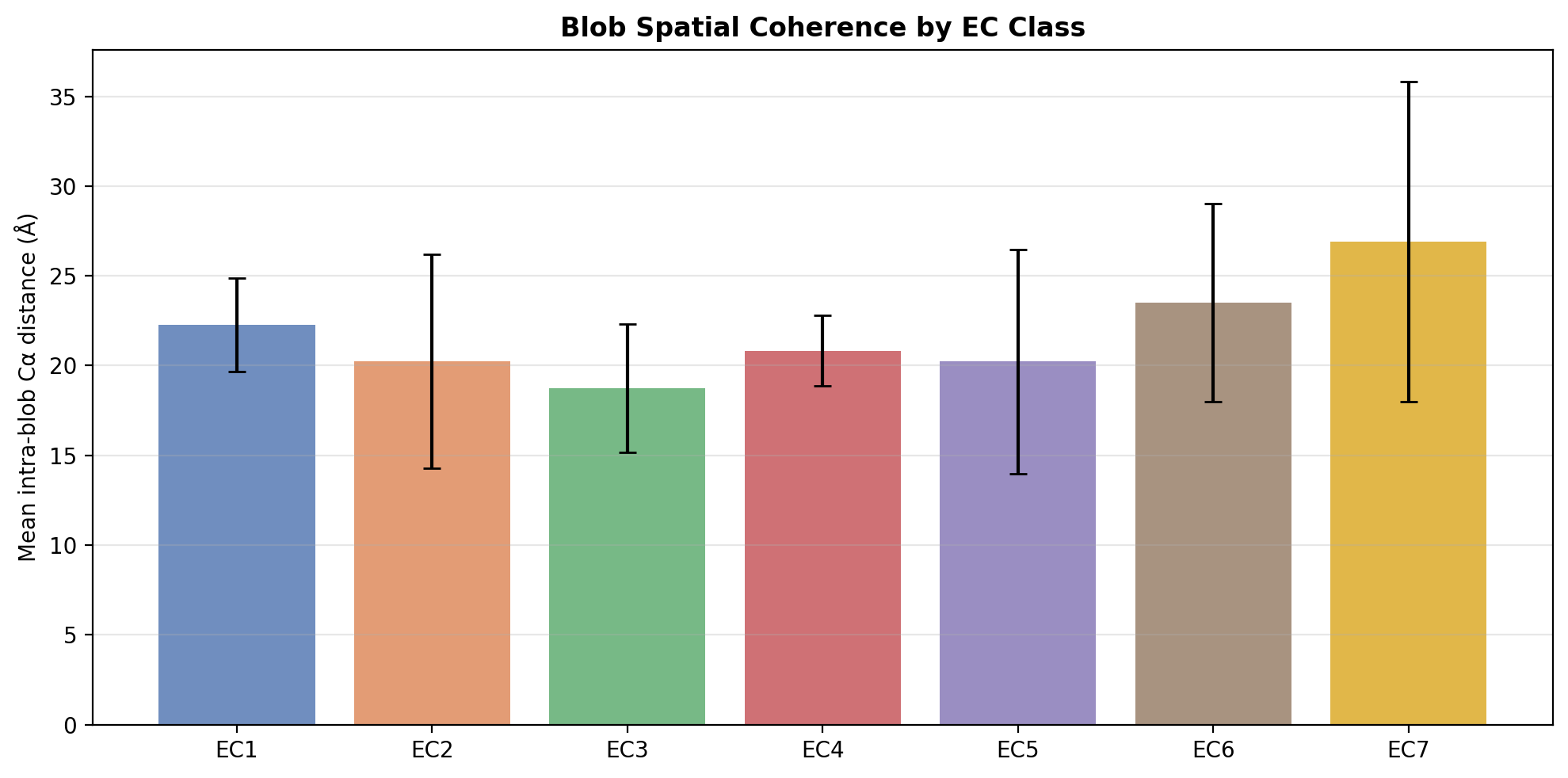}
\caption{Mean intra-blob C$\alpha$ distance across EC classes. Lower
distance indicates greater spatial compactness.}
\label{fig:blob_spatial_coherence}
\end{figure}

\paragraph{Spatial coherence of learned blobs.}
A useful structural decomposition should assign spatially nearby
residues to the same blob. Figure~\ref{fig:blob_spatial_coherence}
shows mean intra-blob C$\alpha$ distances across EC classes.

Learned blobs are spatially compact across all classes, indicating that
assignments are not arbitrary graph partitions.

\begin{figure*}[ht]
\centering
\includegraphics[width=\columnwidth]{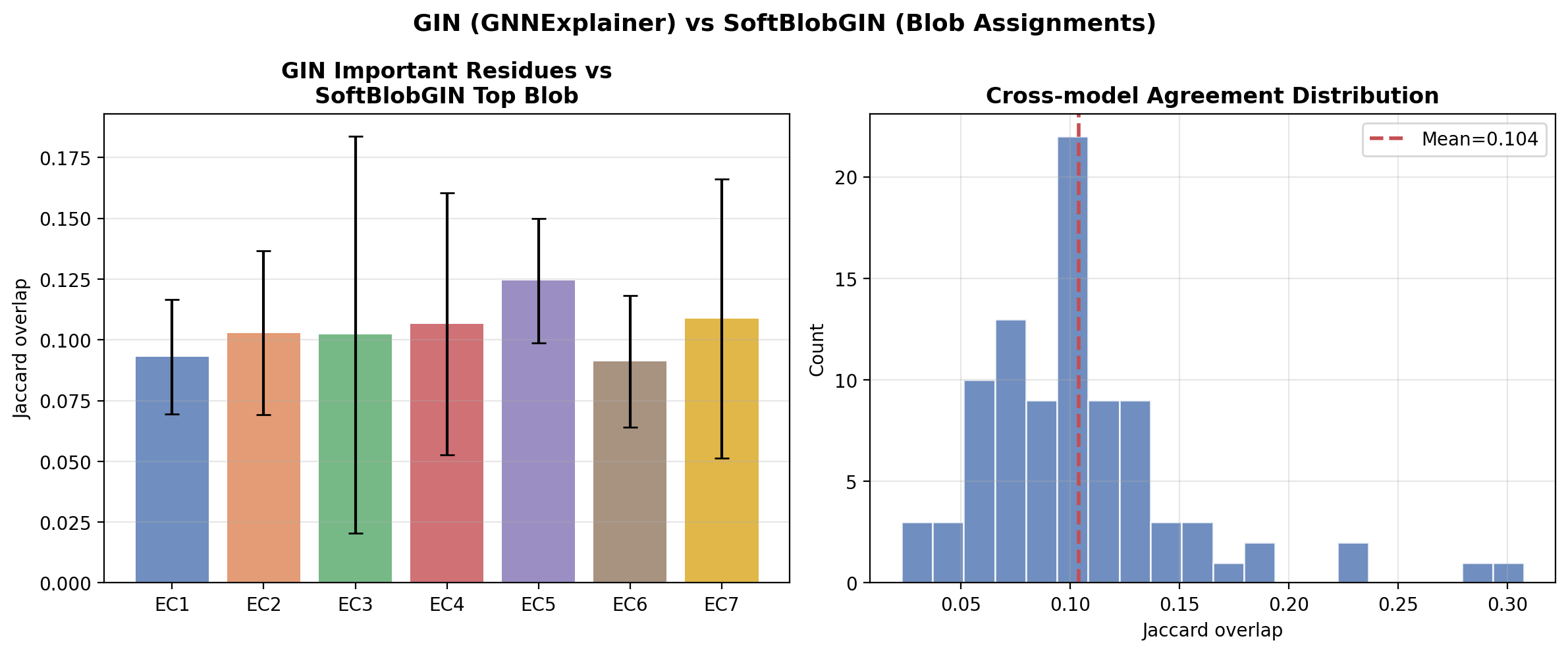}
\caption{Agreement between SoftBlobGIN blob assignments \&
GNNExplainer important residues. Left: per-EC Jaccard overlap. Right:
distribution of overlap scores across proteins.}
\label{fig:gin_blob_overlap}
\end{figure*}

\paragraph{Interpretation of Agreement with post-hoc explainers.}
We compare SoftBlobGIN blob assignments with GNNExplainer edge
saliency. Figure~\ref{fig:gin_blob_overlap} shows modest but non-zero agreement
between top-ranked blobs \& GNNExplainer important residues,
indicating partial overlap between intrinsic \& post-hoc
explanations. Together, solvent accessibility, active-site enrichment, amino acid
composition, spatial coherence, \& agreement with post-hoc explainers
indicate that SoftBlobGIN learns biologically meaningful structural
partitions rather than arbitrary graph clusters.

\begin{figure*}
\centering
\includegraphics[width=0.7\textwidth]{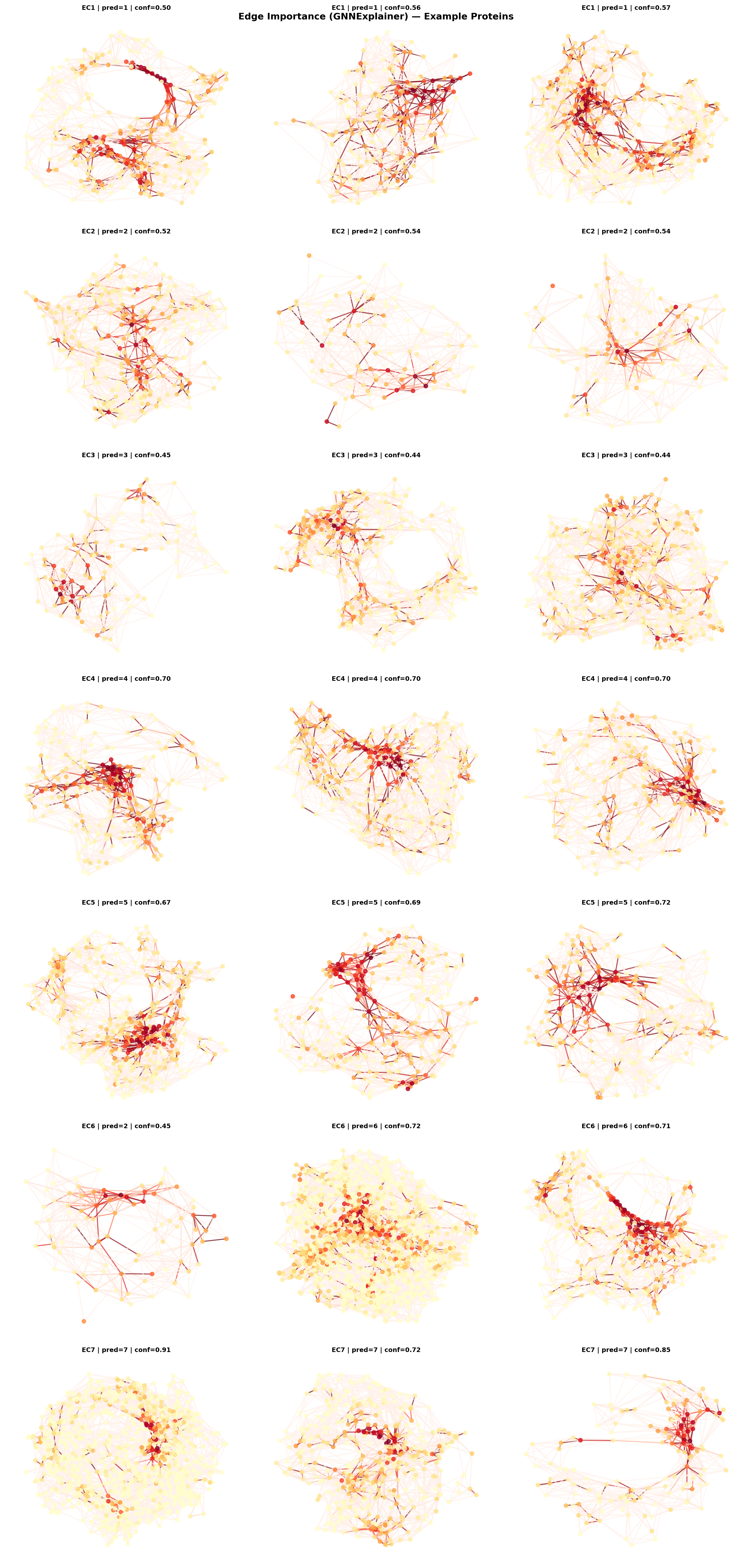}
\caption{Representative GNNExplainer edge-importance maps across EC
classes. Darker edges indicate higher importance.}
\label{fig:gnnexplainer_examples}
\end{figure*}

\section{Relationship to known domain boundaries}
\label{app:domain}

To evaluate whether learned blobs merely reproduce known domain
segmentations, we compare blob assignments against annotated Pfam/CATH
domains. Blob-domain agreement is partial, with mean Jaccard overlap
$0.380$. This indicates that learned blobs are related to known domain
boundaries but do not simply recover canonical domain annotations. 

This behavior is expected: domains represent evolutionary \&
structural units, whereas SoftBlobGIN blobs are optimized directly for
enzyme classification. Learned blobs therefore partially align with
known domains while deviating when functional discrimination benefits
from finer-grained or cross-domain partitions.  Finally, representative visualizations of the GNNExplainer are shown in
Figure~\ref{fig:gnnexplainer_examples}.

\section{Societal Impact}
\label{app:societal_impact}

Our framework is designed to make protein function predictions more
transparent by providing structurally grounded explanations alongside
classifications. This has positive implications for drug discovery,
enzyme engineering, \& clinical genomics, where interpretability
supports regulatory compliance \& scientific trust. We do not foresee
direct negative societal impacts: the method operates on publicly
available protein structures \& does not generate novel sequences or
designs that could be misused. All data used in this work (ProteinShake,
ESM-2 embeddings, PDB/UniProt annotations) are publicly released under
permissive licenses, \& our code \& model checkpoints carry no dual-use
risk beyond standard protein classification tools.

\end{document}